\documentclass[12pt]{article}

\usepackage[utf8]{inputenc}
\usepackage[T1]{fontenc}
\usepackage[margin=1.2in]{geometry}
\usepackage{xurl}
\usepackage{booktabs}
\usepackage{amsfonts}
\usepackage{nicefrac}
\usepackage{microtype}
\usepackage{doi}
\usepackage{amsmath, amssymb}
\usepackage{mathtools}

\usepackage{placeins}
\usepackage{graphicx}
\usepackage{dcolumn}
\usepackage{bm}
\usepackage{hyperref}
\hypersetup{hidelinks}
\usepackage{color}
\usepackage{xcolor}

\usepackage{authblk}
\usepackage[backend=biber]{biblatex}
\title{Generative diffusion models for spatiotemporal influenza forecasting
}
\date{}

\author[1]{Joseph Lemaitre}
\author[1,2,3]{Justin Lessler}
\affil[1]{Department of Epidemiology, Gillings School of Global Public Health, University of North Carolina at Chapel Hill, Chapel Hill, NC 27599, USA}
\affil[2]{Department of Epidemiology, Johns Hopkins Bloomberg School of Public Health, Baltimore, MD 21205, USA}
\affil[3]{Carolina Population Center, University of North Carolina at Chapel Hill, Chapel Hill, NC 27599, USA}

\begin{document}
\pagenumbering{arabic}
\setcounter{page}{1}
\begin{titlepage}
\maketitle
\thispagestyle{plain}

\noindent\textbf{Corresponding author:} Joseph Lemaitre; jo.lemaitresamra@gmail.com; 170 Rosenau Hall, CB \#7400, 135 Dauer Dr. Chapel Hill, NC 27599; +1 (919) 923-7620.

\end{titlepage}
\setcounter{page}{2}

\section*{Abstract}
Forecasting infectious disease incidence can provide important information to guide public health planning, yet it is difficult because epidemic dynamics are complex. Current mechanistic and statistical approaches often struggle to capture multimodal uncertainty or emergent trends. Influpaint adapts denoising diffusion probabilistic models to epidemic forecasting. By encoding influenza seasons as spatiotemporal images in which pixel intensity represents incidence, Influpaint learns a rich distribution of disease dynamics from a hybrid dataset of surveillance and simulated trajectories. Forecasting is formulated as a conditional generation (inpainting) task from partial observations. We show that Influpaint generates realistic, diverse epidemic trajectories and achieves forecast accuracy that is competitive with leading ensemble methods in retrospective evaluation. In real-time evaluation during the 2022--2025 U.S. CDC FluSight challenges, performance improved substantially across seasons, with highly accurate but somewhat overconfident projections in 2024--2025. The best performance was achieved with a training dataset containing 30\% surveillance and 70\% simulated trajectories. These results show that diffusion models can capture important spatiotemporal structure in influenza dynamics and provide a flexible framework for probabilistic infectious disease forecasting.

\section*{Author Summary}
Accurately forecasting influenza activity is difficult, which complicates public health preparedness and response. We propose an influenza forecasting approach, named Influpaint, that treats an influenza season like an image, with weeks on one axis, locations on the other, and color intensity representing reported hospitalizations. This framing lets us use a type of generative artificial intelligence, the diffusion model, to learn the patterns of influenza activity and then predict (inpaint) future weeks from the part of the season already observed. We show that Influpaint generates plausible influenza seasons and produces forecasts that are competitive with established ensemble methods. The approach is also highly flexible and can be used to impute missing data without additional training. We also find that combining real surveillance data with simulated epidemic trajectories improves performance.

\section{Introduction}
Infectious disease forecasting can inform timely public health action, such as resource allocation and interventions. Producing reliable forecasts is challenging. Epidemics are dynamic, complex systems shaped by intrinsic randomness, changing population behaviors, environmental factors, and reporting delays. Accurate forecasting requires models that can handle non-stationary dynamics and limited data availability while capturing the full spectrum of plausible future trajectories.

Most forecasting approaches are based on mechanistic or statistical models. Mechanistic models incorporate epidemiological knowledge, but can generate unrealistic projections when the assumptions fail to describe epidemic dynamics. Statistical time-series methods excel at identifying historical patterns but can be brittle when underlying processes shift and are of limited utility when confronted with a novel pathogen. Forecast uncertainty can involve distinct alternative futures (e.g., continued decline or renewed growth after a midseason downturn)~\cite{osthus_dynamic_2019}, and many mechanistic and statistical models under-represent the multimodality of these predictive distributions.

The new generation of generative AI models includes promising approaches for modeling high-dimensional distributions. In particular, denoising diffusion probabilistic models (DDPMs) have demonstrated paradigm-shifting performance in image and audio synthesis \cite{ho_denoising_2020_final,dhariwal_diffusion_2021_final}. Their ability to produce coherent samples conditioned on partial observations makes them well suited to spatiotemporal epidemic forecasting. While there has been limited work using Generative Adversarial Networks (GANs) and other deep learning methods for epidemic forecasting \cite{ray_flusion_2025,wohlfender_machine_2026,kraemer_artificial_2025}, the capacity of diffusion models to capture nuanced uncertainty and diverse epidemic dynamics remains unexplored.

Here we introduce \textbf{Influpaint}, a generative framework that applies DDPMs to influenza forecasting in the United States. Influpaint represents an influenza season as a two-dimensional image, with time on one axis, location on the other, and pixel intensity reflecting incidence. This representation enables the direct use of modern image-generation techniques, while conditioning on observed history is achieved through an inpainting procedure that enforces coherence between known observations and generated futures \cite{zhang_towards_2023_final}. We describe the Influpaint architecture and training datasets, evaluate its ability to reproduce realistic epidemic dynamics, and assess both its retrospective forecasting skill and its performance over the past three FluSight seasons (2022--2025). We benchmark alternative model formulations and quantify the contributions of key design choices such as dataset composition and inpainting schedule.

\section{Method overview}
Influpaint represents an influenza season as a matrix with weeks and locations as axes and incident influenza hospitalizations as values. It is trained on a collection of past seasons obtained from surveillance data and model simulations. Influpaint can then generate new seasons by progressively denoising random noise, which we call \textit{unconditional generation}. For forecasting, we condition this generation process on observed hospitalizations using \textit{inpainting} and provide a \textit{mask} that differentiates the observed entries from those to be generated. The model generates future weeks while preserving known observations. Further details are provided in \textbf{Materials and Methods}.
\section{Results}

\subsection{Unconditional generation of realistic influenza seasons}
\begin{figure} \centering

	\includegraphics[width=.75\textwidth]{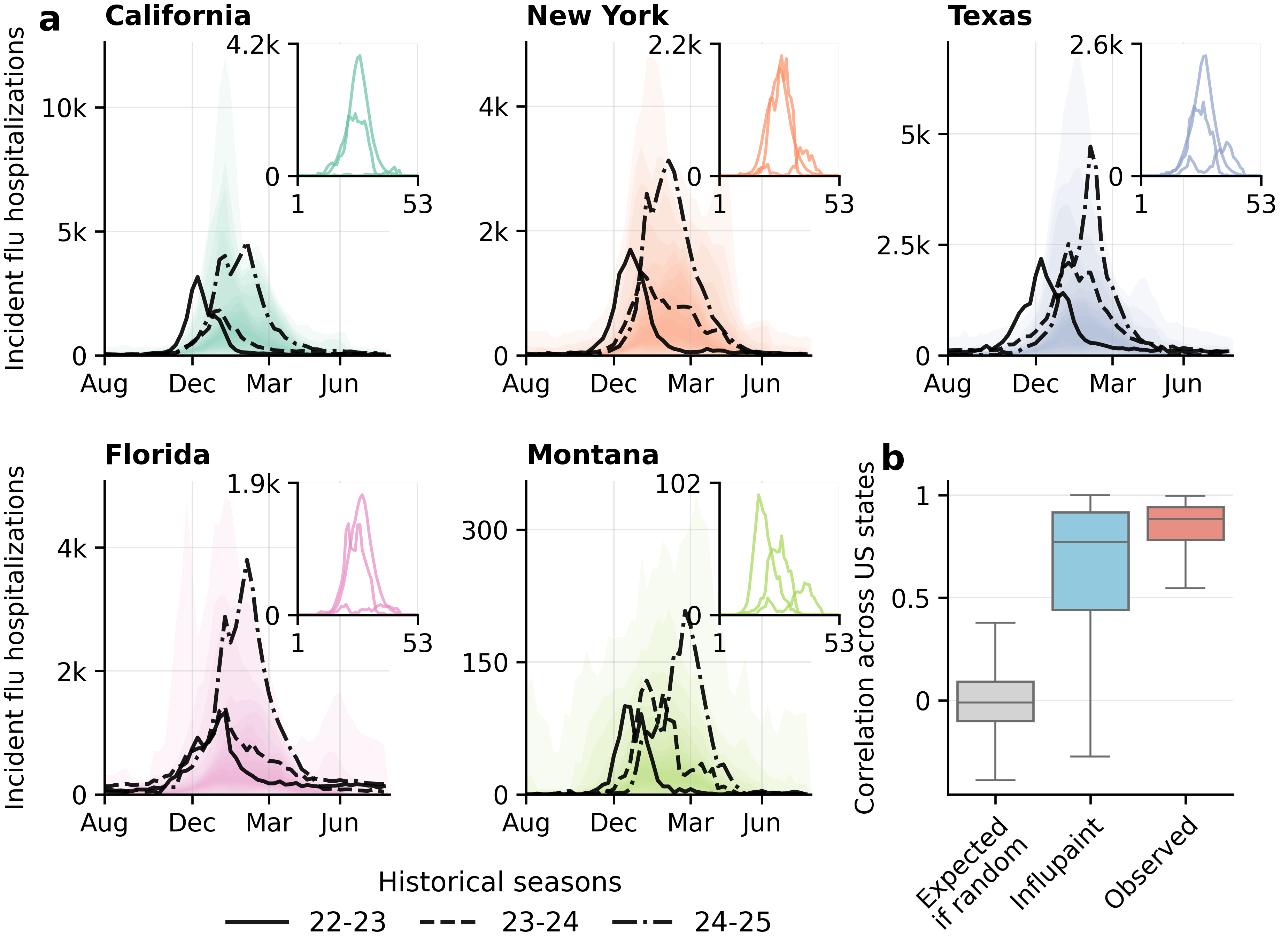}
	\caption{
\textbf{Influenza trajectories generated by Influpaint.}
\textbf{a.} The colored bands show the quantiles across 512 generated trajectories of weekly hospital admissions for California, New York, Texas, Florida, and Montana; three example trajectories are highlighted in the inset. Black lines show realized historical influenza seasons from the National Healthcare Safety Network data for each state.
\textbf{b.} Pairwise cross-state weekly incidence correlations from Influpaint-generated trajectories, compared with a time-permuted null expectation and with observed correlations from recent historical seasons.
}
\label{fig:uncond_gen}\end{figure}

When generating unconditional trajectories (i.e., seasonal trajectories with no observed data), Influpaint produces novel (non-training) samples that respect known influenza epidemic patterns, including varying periods of increase and decline with clear seasonal peaks (Fig~\ref{fig:uncond_gen}a). This indicates that the model has learned a rich, high-dimensional representation of influenza dynamics. Visual examination of synthetic seasons suggests a high degree of realism and diversity.
For instance, Influpaint generates seasons with a single high-intensity peak as well as seasons with bimodal epidemic curves, two patterns previously observed (e.g., in 2023--2024 and 2024--2025). This variety is difficult to replicate with classic epidemic models. The model further produces seasons with unusually early or late peaks, as well as seasons with low, geographically scattered incidence. The envelope produced by 512 trajectories contains the range of incidence seen in recently observed seasons (2022--2023, 2023--2024, and 2024--2025).

Likewise, Influpaint recovers some of the qualitative structure of multi-state epidemic timing (Fig~\ref{fig:uncond_gen}b), with projections showing stronger spatial synchrony than expected by chance (mean inter-state correlation $0.646$), though still below the levels observed in the 2023--2024 and 2024--2025 influenza seasons (mean inter-state correlation $0.833$).

\subsection{Forecasting via temporal inpainting}
We next evaluate Influpaint's core application: forecasting future trajectories conditioned on partially observed seasons. We evaluate Influpaint's retrospective performance over two FluSight seasons: 2023--2024 and 2024--2025.

\begin{figure} \centering

	\includegraphics[width=\textwidth]{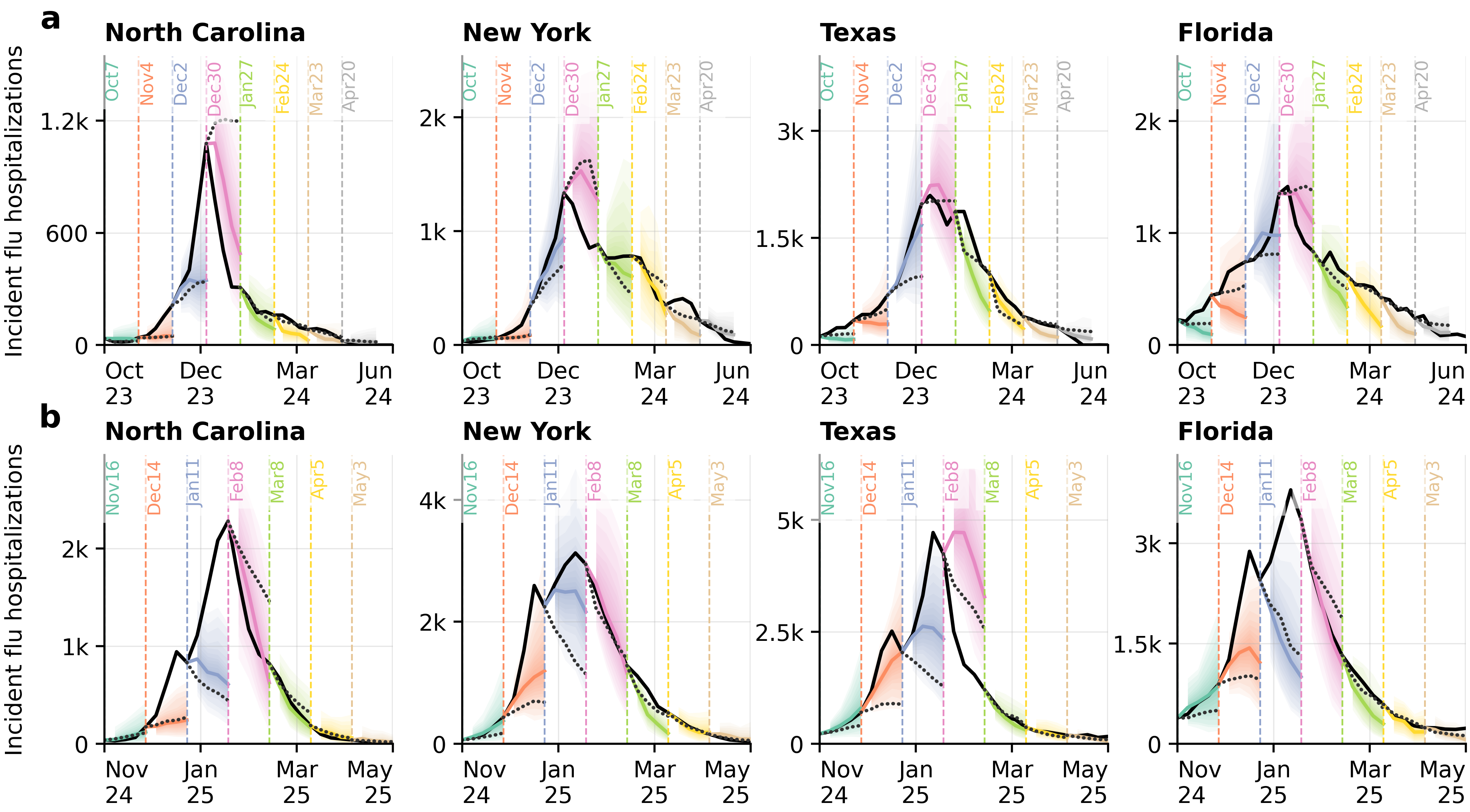}
\caption{
\textbf{4-week-ahead forecasts.}
\textbf{a.} Each panel shows 4-week-ahead forecasts for North Carolina, New York, Texas, and Florida at multiple forecast dates. The dashed colored vertical lines mark the last week of observed data for a given forecast. Forecast uncertainty is summarized by quantiles (colored fan) and the median (colored line) from 512 conditional trajectories. The solid black line shows the observed final values. For the same reference dates, the FluSight ensemble forecast is shown as a dotted line.
\textbf{b.} Same as panel a, but for the 2024--2025 influenza season.
}
\label{fig:forecast_fourweek}\end{figure}

Fig~\ref{fig:forecast_fourweek} shows forecasts at a four-week horizon, which is frequently used in infectious disease forecasting competitions, alongside projections from the FluSight ensemble. The ensemble aggregates submitted forecasts by taking the median across model submissions at each predictive quantile. It is commonly regarded as among the best and most consistent available forecasting models. Because Influpaint does not model reporting delay, this retrospective evaluation uses finalized, revised values for the observed weeks up to each forecast date. No observations from weeks after the forecast date are used to condition the forecasts. These inputs differ from the unrevised data available to participating models in real time. \textbf{Section 3.4} reports Influpaint's operational performance using data available in real time.

Influpaint forecasts often anticipate turning points, with trajectories steepening and then drifting downward near seasonal maxima. Quantitatively, Influpaint performed competitively with the CDC FluSight multi-model ensemble across the two seasons. We evaluate probabilistic accuracy using the Weighted Interval Score (WIS), a proper scoring rule commonly used in influenza forecasting, for which lower values indicate better performance \cite{Bracher:EvaluatingEpidemicForecasts:2021}. We compare Influpaint against models that submitted to FluSight and missed five or fewer forecasts: 32 models in 2023--2024 and 42 in 2024--2025. Using finalized ground-truth data, Influpaint achieved a total WIS of 185{,}441 in 2023--2024 (ranked 5\textsuperscript{th} of 33) and 450{,}797 in 2024--2025 (ranked 8\textsuperscript{th} of 43), compared with 197{,}834 and 554{,}066 for the FluSight ensemble (ranked 8\textsuperscript{th} and 20\textsuperscript{th}, respectively). Coverage was well calibrated, though slightly overconfident: empirical 50\% and 90\% interval coverages were 0.48 and 0.85 in 2023--2024 and 0.51 and 0.85 in 2024--2025, compared with 0.50 and 0.86 for the ensemble in 2023--2024 and 0.50 and 0.76 in 2024--2025. Although this comparison is biased because Influpaint forecasts were generated retrospectively using finalized values for observations up to each forecast date rather than under real-time reporting constraints, it nonetheless shows that a diffusion-based generative model can achieve accuracy and calibration comparable to those of leading operational ensembles while offering coherent, sample-based representations of uncertainty and diverse epidemic trajectories.

\begin{figure} \centering

	\includegraphics[width=\textwidth]{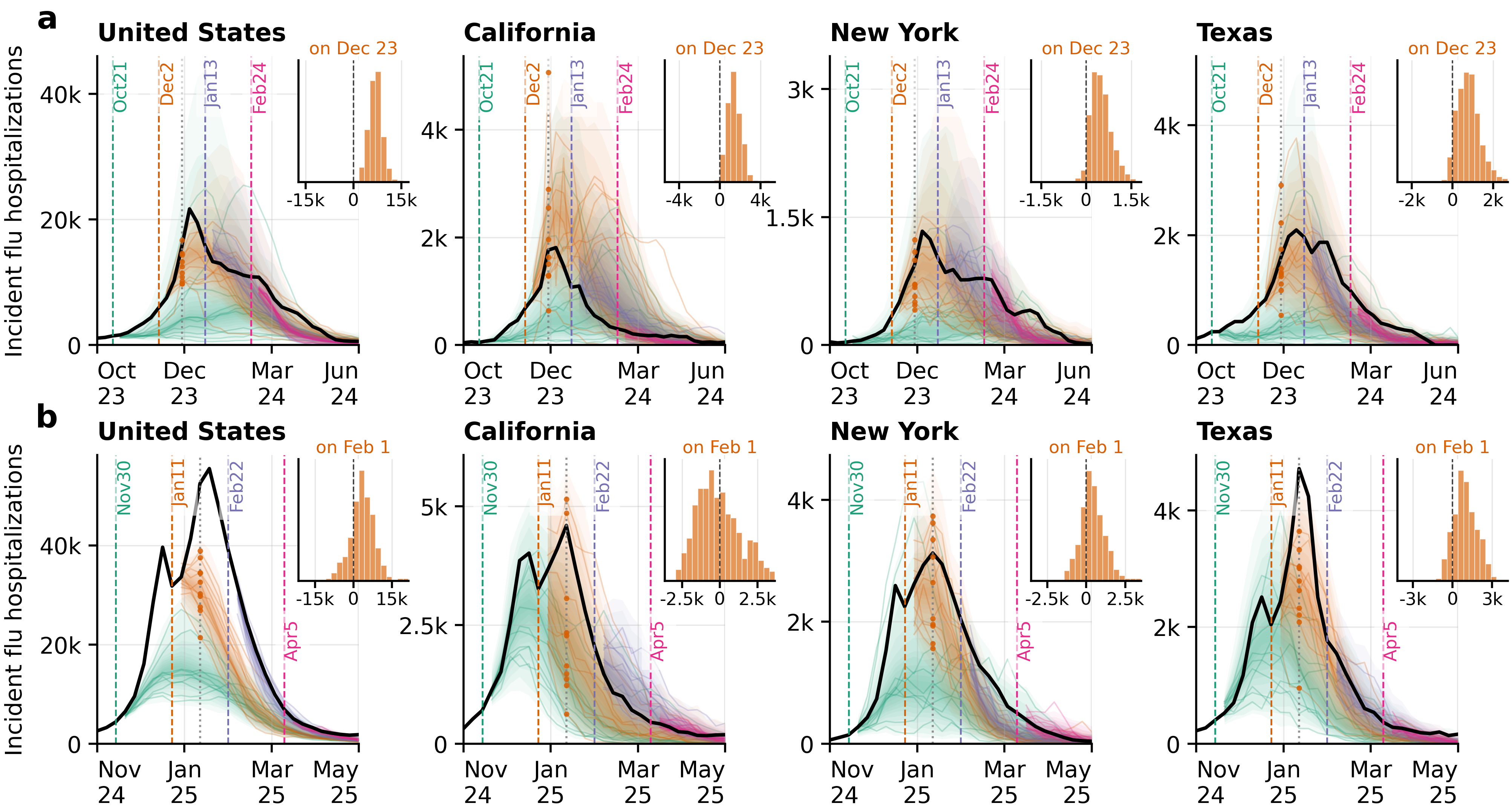}

\caption{\textbf{Full-season forecasts.}
\textbf{a.} Each panel for the United States, California, New York, and Texas shows reported hospitalizations (black line). The colored dashed vertical line marks the last observed week for a given forecast, with the forecast quantiles from 512 conditional trajectories (colored bands) representing the probabilistic forecast. Ten example trajectories are highlighted as thin colored lines. The histograms show changes in hospitalizations across the 512 orange forecast trajectories from December 2 to December 23, 2023, in panel \textbf{a} and from January 11 to February 1, 2025, in panel \textbf{b}. The gray dotted lines mark these target weeks. Negative and positive changes indicate declining and increasing hospitalizations, respectively, with zero change marked by a dashed line in each inset.
\textbf{b.} Same as panel a, but for the 2024--2025 influenza season.}
\label{fig:forecast_tillend}\end{figure}

Fig~\ref{fig:forecast_tillend} extends the forecast horizon to the end of the season. Early in the season, the predictive distribution is multimodal, reflecting divergent but plausible seasonal trajectories. The histogram insets in Fig~\ref{fig:forecast_tillend} illustrate this divergence, with some trajectories predicting increases in hospitalizations and others predicting declines over a fixed three-week period. As intended, uncertainty is tight near the beginning of the forecast and retains reasonable empirical coverage as the horizon grows. Qualitatively, Influpaint preserves informative seasonal signal at long horizons while adapting quickly as weekly observations reveal the dominant trajectory. For longer horizons, performance predictably degrades relative to short-term ensemble forecasts.

\subsection{Training data mix and model formulation determine forecast skill}
Having established forecast performance, we next quantified which design choices were most responsible for Influpaint's skill. We ran one-at-a-time ablations on 1--4-week-ahead state forecasts in 2023--2024 and 2024--2025 and evaluated each variant.

The composition of the training data had an important effect. We compared four training sets: surveillance only (CDC FluView and FluSurv), simulation only (Flu Scenario Modeling Hub and flepiMoP trajectories \cite{lemaitre_flepimop_2024, loo_us_2024}), and two hybrids (30\%/70\% and 70\%/30\% surveillance/simulation). Both hybrids outperformed single-source training, and the 30\% surveillance / 70\% simulation mix gave the best WIS profile across seasons and horizons (Table~\ref{tab:dataset_mix_effect}). Despite its larger size ($1{,}240$ unique training samples, compared with $20$ for the surveillance-only dataset), the simulation-only dataset produced worse performance.

\begin{table}[t]
\centering
\setlength{\tabcolsep}{5pt}
\begin{tabular}{lrrr}
\hline
Dataset composition & $N_{samples}$ & Total WIS & Relative performance (\%) \\
\hline
30\% surveillance / 70\% modeling & $1,260$ & $647,816$ & baseline \\
70\% surveillance / 30\% modeling & $1,260$ & $704,939$ & $-8.82$ \\
100\% surveillance & $20$ &$  707,704$ & $-9.24$ \\
100\% modeling & $1,240$ & $724,125$ & $-11.78$ \\
\hline
\end{tabular}
\caption{\textbf{Effect of training-data composition on forecast skill.} Each row reports the WIS over the 2023--2024 and 2024--2025 seasons for the single model obtained by changing only the dataset composition and keeping all other settings at their baseline values. Improvements are reported relative to the baseline 30\% surveillance / 70\% modeling mix, matching the dataset-composition entries in Fig~A in S1 Appendix. Positive values indicate better performance than the baseline; negative values indicate worse performance. $N_{samples}$ is the number of unique source trajectories before repetition. Both hybrid datasets contain 1,240 simulated and 20 surveillance samples. The mixing ratio is achieved by repeatedly sampling surveillance frames to obtain the desired proportion in the training dataset.}
\label{tab:dataset_mix_effect}
\end{table}

Other formulation choices had measurable effects. Notably, using 500 denoising steps rather than 200 and using square-root scaling both improved forecasting performance. Additional training perturbations intended to enrich the training dataset (Poisson resampling, temporal padding, intensity scaling) generally worsened performance. By contrast, alternative U-Net variants and inpainting schedules produced only small differences (Fig~A in S1 Appendix).

\subsection{Realized performance in the FluSight challenge}
To evaluate Influpaint in an operational setting, we submitted its forecasts under the team name \texttt{UNC\_IDD-InfluPaint} to the U.S. CDC FluSight forecasting challenge for the 2022--2023, 2023--2024, and 2024--2025 influenza seasons. FluSight is conducted in real time: each week, participating models submit probabilistic forecasts of influenza hospital admissions at the national and state levels, which are evaluated once observations become available. Across the three seasons in which we submitted forecasts, Influpaint's performance improved markedly as the algorithm matured, placing it among the top models in the most recent season (2024--2025).

Our first FluSight participation, for the 2022--2023 season, used an early Influpaint implementation based on the RePaint inpainting algorithm. This version exhibited boundary misalignment between observed and generated weeks, which degraded forecast calibration. The model ranked 9\textsuperscript{th} of 17 participating systems, with 50\% prediction-interval coverage of 0.36~\cite{mathis_evaluation_2023_final} (9\textsuperscript{th} of 17 on mean absolute error (MAE) and 10\textsuperscript{th} of 17 on relative WIS, which compares the model with a constant baseline).

Across the two most recent FluSight seasons, Influpaint showed marked improvement as the framework matured and the training data and conditioning algorithms were refined. Among models submitting at least 70\% of weekly forecasts (excluding FluSight ensemble variants), \texttt{UNC\_IDD-InfluPaint} ranked near the bottom of the leaderboard in 2023--2024 (absolute WIS = 61.9, rank 24/27; relative WIS = 1.10, rank 20/27; MAE = 87.8, rank 24/27), with empirical 50\% and 90\% coverages of 28.7\% and 71.7\%. In contrast, in 2024--2025 the model achieved substantial gains in sharpness and accuracy, ranking 1\textsuperscript{st} of 36 models for both absolute WIS (97.9) and MAE (128.3), and 11\textsuperscript{th} for relative WIS (0.71). However, coverage was substantially below target (14.8\% at 50\%, 52.6\% at 90\%), indicating that while forecasts became sharper, uncertainty was underestimated.

\subsection{Generalization to heterogeneous masking patterns}
Because conditioning is implemented through inpainting, Influpaint can enforce arbitrary observation masks at sampling time: any subset of spatiotemporal entries (weeks $\times$ locations) can be fixed to observed values, and the remaining entries are sampled from the learned conditional distribution. This makes the approach well matched to temporal reporting gaps and partial spatial coverage.

\begin{figure} \centering

\includegraphics[width=\textwidth]{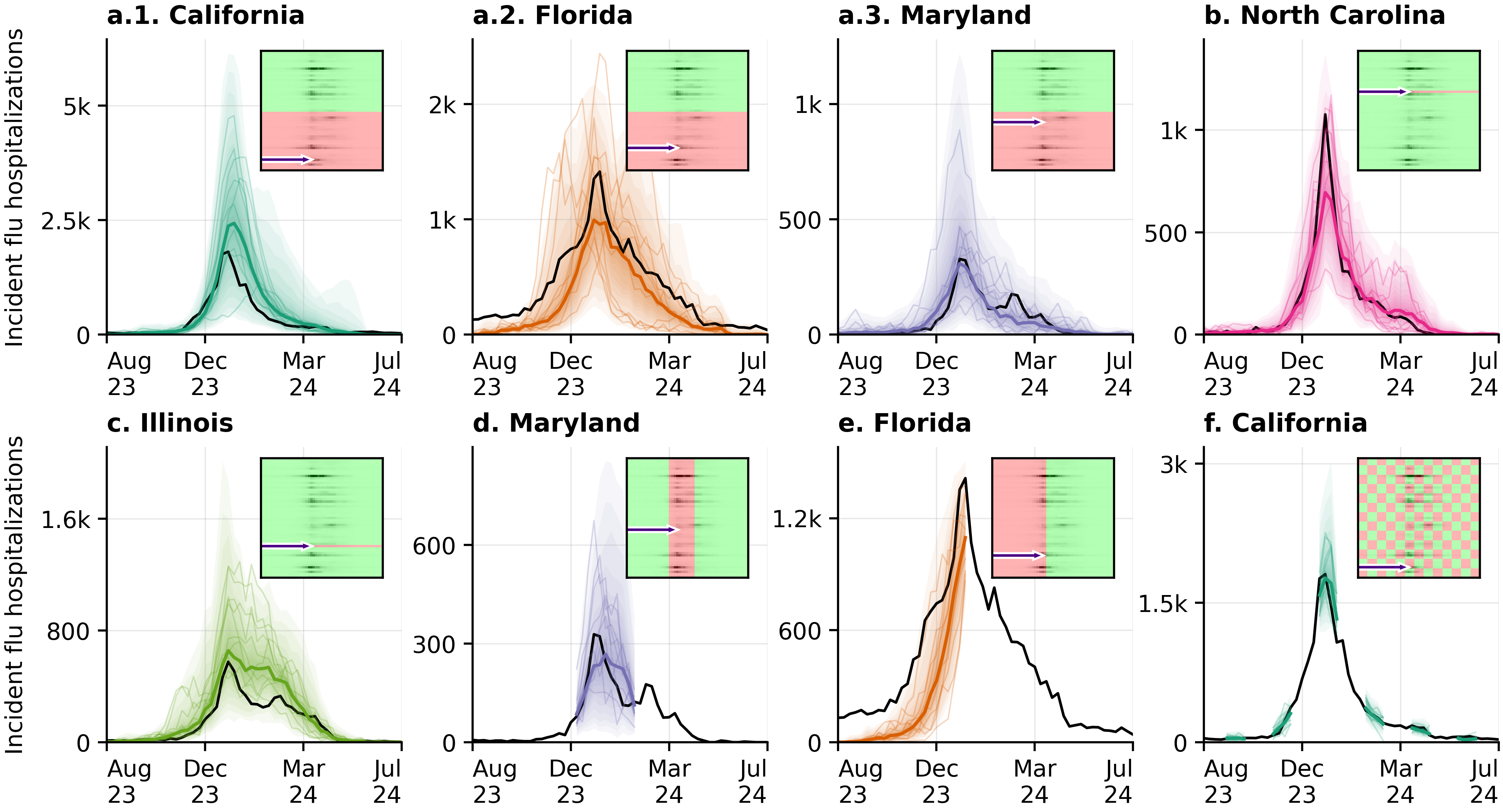}
\caption{
\textbf{Robustness to mask design.} Black curves show observed hospitalizations during the 2023--2024 season; colored fans and lines show Influpaint predictive quantiles and medians. Insets show the conditioning mask for each panel, where green values are observed and red values are hidden and reconstructed by Influpaint. The arrow in each inset indicates which state/row is shown in the graph.
\textbf{a.1--3.} Half-subpopulation spatial mask for California (a.1), Florida (a.2), and Maryland (a.3).
\textbf{b.} Leave-one-state-out mask for North Carolina.
\textbf{c.} Leave-one-state-out mask for Illinois.
\textbf{d.} Midseason gap mask for Maryland.
\textbf{e.} Reconstruction of the early-season dynamics conditioned on later-season observations for Florida.
\textbf{f.} Checkerboard spatiotemporal mask for California.
}
\label{fig:masks}\end{figure}

Empirically, Influpaint handles partial information well and reconstructs coherent, plausible trajectories across diverse mask designs (Fig~\ref{fig:masks}). For spatial reconstruction, Influpaint preserves broad peak-timing and peak-magnitude structure when a mask withholds 50\% of locations (Fig~\ref{fig:masks}a.1--a.3), and infers plausible trajectories for fully unobserved states from surrounding context under leave-one-state-out masks (Fig~\ref{fig:masks}b,c). Temporal reconstruction is also informative: when only part of a season is observed, the model recovers plausible peak timing and intensity with wider uncertainty, reflecting the limited constraints (Fig~\ref{fig:masks}d). As expected, reconstruction degrades when local dynamics deviate strongly from neighboring-state patterns. In Florida in 2023--2024, the early-season incidence was exceptionally high compared with that in neighboring states. Influpaint fails to capture this elevated incidence under future-only conditioning (Fig~\ref{fig:masks}e). Finally, we demonstrate Influpaint's flexibility with a checkerboard spatiotemporal mask (4 weeks $\times$ 4 states; Fig~\ref{fig:masks}f).

\FloatBarrier

\section{Discussion}
This study shows how diffusion models can be used for infectious disease forecasting. By encoding each influenza season as a two-dimensional time-by-location frame, Influpaint learns a joint spatiotemporal distribution and can generate entire seasons that are realistic in both timing and spatial dynamics. From DDPMs, it inherits a flexible inductive bias and the ability to span a diverse set of plausible outcomes, which is central to expressing uncertainty in forecasts. To our knowledge, this is the first application of DDPMs to infectious disease forecasting.

Influpaint achieved competitive performance in the CDC FluSight challenge, even in years perturbed by the COVID-19 pandemic, and performance improved substantially across three seasons as the framework matured. While our method did not outperform many existing approaches, which often incorporate decades of epidemiological expertise, we show that a generative model based purely on deep learning can produce competitive real-time forecasts. Moreover, across FluSight seasons, no single method consistently outperforms alternatives across all targets and epidemiological contexts. Achieving this level of performance provides a proof of concept for the potential of diffusion models. Further, there are clear pathways to improve performance, such as training on the forecast task more directly and including nowcasting in the forecasting pipeline.

We identified dataset composition as a major determinant of forecast performance: models trained on hybrid datasets containing both surveillance and simulation data consistently outperformed models trained on single-source datasets, with the 30\%/70\% surveillance/simulation mix giving the strongest forecasting performance. In other fields, such as weather forecasting, synthetic datasets have been central to the development of deep learning methods \cite{lam_learning_2023, price_probabilistic_2025}. Here, Influpaint leverages the rich collection of influenza models assembled by the Flu Scenario Modeling Hub \cite{loo_us_2024}.

Finally, Influpaint can generate conditional samples under arbitrary temporal and spatial observation patterns. This flexibility is inherited from DDPM inpainting methods \cite{rout_theoretical_2023} and allows the same trained model to support a variety of tasks relevant to public health, such as interpolating missing weeks and/or locations. Moreover, Influpaint produces coherent trajectory ensembles, so downstream quantities such as peak timing, peak magnitude, and cumulative burden are obtained directly from sampled predictive distributions. The same architecture can be deployed across geographic scales when compatible data are available, enabling a unified workflow from national to subnational forecasting.

Like other forecasting studies, the present work has two important limitations: it is difficult to assess the real-time performance of the final optimized algorithm across multiple seasons, and we cannot fully disentangle limitations specific to this implementation from limitations of the broader diffusion-based forecasting framework. In addition, several Influpaint-specific factors limit its widespread utility in its current form. Performance remains bounded by the diversity and realism of the training data. The hybrid dataset mitigates data scarcity but carries biases from both surveillance and modeling sources. Our ablations suggest that simulated trajectories are an important component for deep learning forecasts, as in other domains. At the same time, simulation-only training underperformed surveillance-only training despite the larger number of samples, raising concerns about the quality and diversity of available synthetic data and their ability to capture the range of epidemic structures. Influpaint also does not yet handle reporting delays or backfill correction, which may affect real-time applications in which data are revised over time. Moreover, interpretability remains a challenge: as a deep generative model, Influpaint does not provide mechanistic explanations for its predictions. Finally, computational cost is substantial: training requires several hours on high-end GPUs, and producing full ensembles of trajectories is slower than for many other models.

Future extensions could improve forecasting performance, at the expense of flexibility, by training conditional diffusion models directly for forecasting rather than conditioning during generation. The current framework could also be expanded by including auxiliary channels such as climate, mobility, or virological data to inform dynamics through additional covariates. Another promising area is the combination of Influpaint with mechanistic models, either by using mechanistic outputs as additional covariates or by using diffusion models to learn residual processes around mechanistic forecasts. Beyond influenza, the Influpaint framework is general and could be extended to other pathogens with rich spatiotemporal data, including RSV, COVID-19, and dengue.

Influpaint demonstrates that recent developments in generative AI, and diffusion models in particular, can deliver competitive and flexible infectious disease forecasts. As surveillance and simulation data ecosystems continue to improve, generative diffusion models may become an important component of epidemic forecasting systems, providing a flexible method for anticipating and understanding disease dynamics.
\FloatBarrier

\section{Materials and Methods}
\paragraph{Overview}
The Influpaint workflow has three steps. First, epidemic trajectories from surveillance and synthetic influenza data are encoded into image-like representations. Second, these representations are used to train a generative model. Finally, generation is conditioned on past observations to produce probabilistic forecasts.

\begin{figure}
\centering
\includegraphics[width=\textwidth]{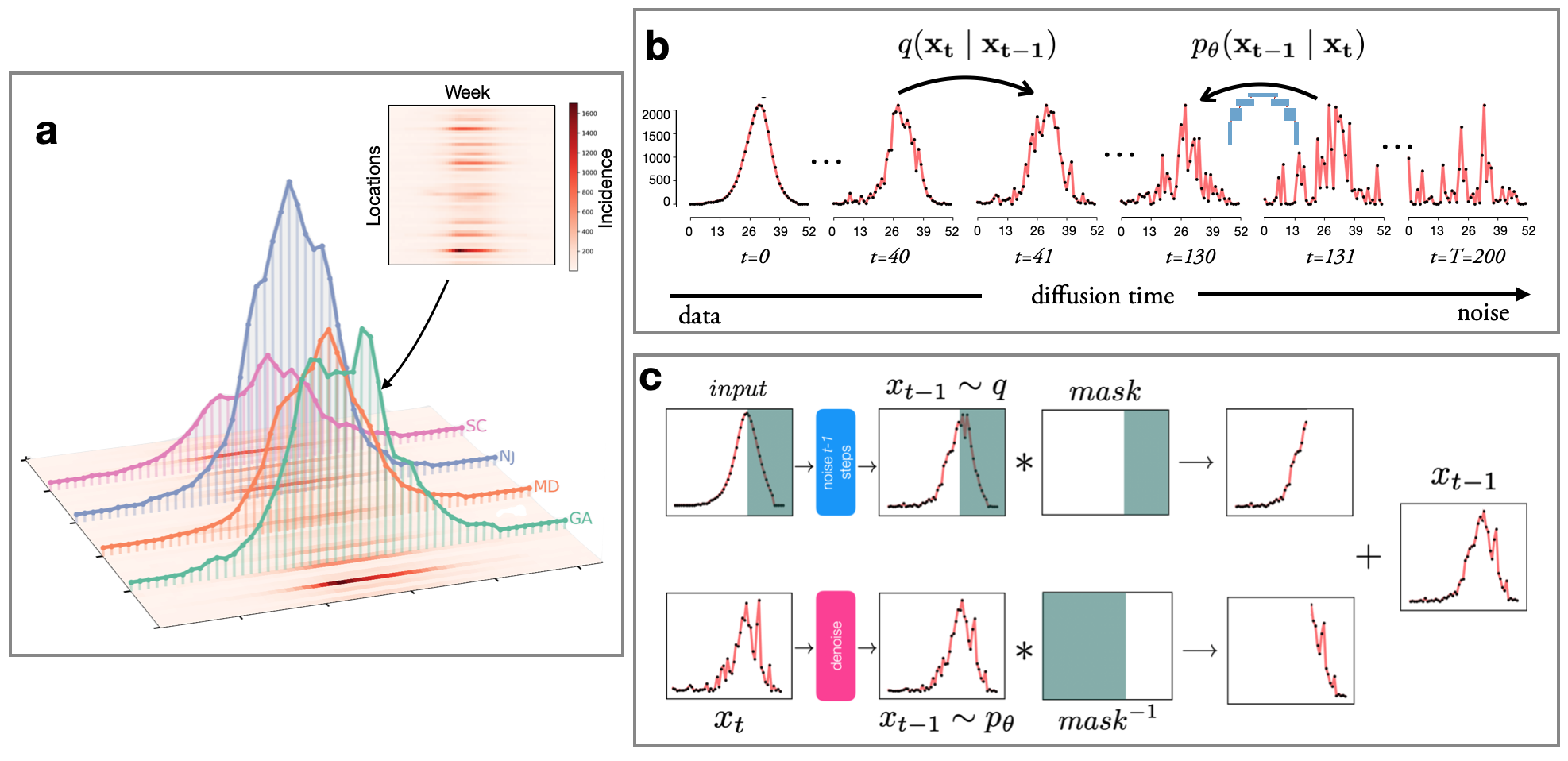}
\caption{\textbf{Methods behind Influpaint.}
	\textbf{a.} Epidemic seasons are represented as images with time and space as axes, and pixel intensity corresponding to incident hospitalizations.
	 \textbf{b.} Denoising Diffusion Probabilistic Models generate data by training a U-Net to restore an image that has been progressively corrupted by Gaussian noise.
	 \textbf{c.} RePaint algorithm for forecasting. The ground truth and a mask (white: values to preserve; green: values to discard and replace by the generation process) are provided. At each denoising step, we stitch the noised ground truth and the intermediate denoised sample to obtain the updated trajectory.
	 }
\label{fig:methods}
\end{figure}

\subsection{Epidemic trajectories as images}
Influenza seasons are represented as two-dimensional arrays (or \textit{frames}) $X$ of shape $52 \times L$, where the $x$-axis corresponds to time in weeks and the $y$-axis to locations (Fig~\ref{fig:methods}a). Each element $X_{ij}$ denotes a reported epidemiological quantity (e.g., incident influenza hospitalizations) in week $i$ for location $j$. In the U.S. influenza context, $L=51$ represents all 50 states plus D.C.

This encoding is analogous to a grayscale image, with pixel intensity proportional to incidence, which enables the direct use of image-generation models. Although the temporal and spatial axes do not have the same semantics, and human mobility or spatial dependence may not be fully captured by convolution kernels, we find that diffusion-model architectures generalize effectively to epidemic generation. Moreover, we found that the ordering of states along the location axis did not affect Influpaint's performance (Kupperman et al., using a different model, report that ordering influenced the performance of genetic HIV outbreak detection \cite{kupperman_deep_2022}). Influpaint's robustness may be explained by the flexibility of diffusion models, although we believe that incorporating inductive biases tailored to spatial epidemic dynamics could improve performance.

\subsection{Training data}
Diffusion models require a large number of high-quality samples for training, but historical influenza datasets are limited, incomplete, and heterogeneous.
To provide sufficient training data, we built a hybrid dataset combining reported hospitalization data and synthetic epidemic trajectories generated by mechanistic influenza models. It features:
\begin{itemize}
    \item CDC FluView (via the Delphi Epidata API; percent influenza-like illness).
    \item CDC FluSurv-NET hospitalizations (via the Delphi Epidata API; extended to all states by population weighting).
    \item Synthetic influenza trajectories from multiple modeling teams, as archived in the Flu Scenario Modeling Hub:
        \begin{itemize}
            \item Round 4 (2023--2024): 4 models, 600 trajectories each across 6 scenarios.
            \item Round 5 (2024--2025): 7 models, at least 600 trajectories each across 6 scenarios.
            \item Round 1 (2022--2023): flepiMoP model, 1,200 trajectories across 4 scenarios.
        \end{itemize}
\end{itemize}

\paragraph{Frame library.} To make all frames complete seasons (all locations and weeks), we assembled seasons from available sources while preserving cross-state relationships and realistic temporal structure. To avoid oversampling the synthetic dataset, we selected 20 trajectories per scenario per model. The final frame library contains 1,080 frames from the Flu Scenario Modeling Hub rounds 4 and 5 and 160 from flepiMoP round 1, in addition to the 20 surveillance frames (13 years of FluView and 7 years of FluSurv).

\paragraph{Dataset compositions.} From this frame library, we constructed four datasets with varying surveillance/modeling mixes:
\begin{itemize}
    \item 100\% surveillance (20 frames, repeated to form a dataset of size 520, larger than our batch size).
    \item 100\% modeling (1,240 frames).
    \item 30\% surveillance / 70\% modeling (1,260 unique frames, repeated to 3,000).
    \item 70\% surveillance / 30\% modeling (1,260 unique frames, repeated to 3,000).
\end{itemize}

\paragraph{Transforms and augmentations.} To harmonize data sources, FluView percentages of influenza-like illness were rescaled using the distribution of the Flu Scenario Modeling Hub projected peak intensities. We applied square-root scaling before rescaling all frames to the $[0,2]$ range ($z = 2 \sqrt{x/M}$, where $z$ is the transformed version of a dataset sample $x$, and $M$ is the dataset maximum). Inspired by image-generation augmentation methods \cite{krizhevsky_imagenet_2012}, we explored targeted augmentations designed to enrich variability while preserving realism: Poisson resampling (observational noise), temporal padding ($\pm4$ or $\pm15$ weeks), and intensity rescaling by a factor $\alpha$ with either narrow ($0.7$--$1.3$) or wide ($0.1$--$1.9$) ranges.

We therefore evaluated four dataset compositions, two transforms (linear vs square-root), and four enrichment schemes (none; Poisson only; Poisson + narrow pad/scale; Poisson + wide pad/scale).

As detailed in S1 Appendix, performance was best without augmentation on the 30\% surveillance / 70\% modeling dataset.

\subsection{Generative model: denoising diffusion probabilistic model}
A Denoising Diffusion Probabilistic Model (DDPM) consists of two processes. The forward (noising) process repeatedly adds Gaussian perturbations, progressively destroying the signal until only noise remains after $T$ steps. The backward process applies a neural network trained to remove that noise step by step, so generation can start from pure noise and produce new, realistic seasons.

DDPMs were introduced for image generation by Ho et al. \cite{ho_denoising_2020_final} and extended by Dhariwal and Nichol \cite{dhariwal_diffusion_2021_final}. Below, we summarize the governing principles as used in Influpaint and refer the reader to these works for the full derivations.

\paragraph{Forward (noising) process.} We start by drawing a trajectory from our data distribution $x_0 \sim q(x_0)$ and repeatedly perturb it with Gaussian noise according to a variance schedule $\beta_1, \ldots, \beta_T$. The variance schedule is defined so that after all forward steps, when $t=T$, the trajectory is indistinguishable from white noise, as shown in Fig~\ref{fig:methods}b. After $t$ steps of independent noising, the result can be expressed in closed form as a Gaussian:
\begin{equation}
    q(\boldsymbol x_t \mid \boldsymbol x_0) = \mathcal{N}\!\left(\boldsymbol x_t; \sqrt{\bar\alpha_t}\, \boldsymbol x_0, (1 - \bar \alpha_t) \mathbf{I} \right), \quad \text{where~}   \bar \alpha_t = \prod_{s=1}^t (1 - \beta_s).
\label{eq:forward}
\end{equation}

The mean is the original trajectory scaled by $\sqrt{\bar\alpha_t}$, and the variance term accumulates the injected noise. Note that while the training data lie in the $[0,2]$ range, intermediate diffusion steps are not restricted to this interval.

\paragraph{Reverse (denoising) process.} Generation starts by drawing from a pure Gaussian prior, $p(\boldsymbol x_T) = \mathcal{N}\left(\boldsymbol x_T; \mathbf{0}, \mathbf{I}\right)$. A neural network $\boldsymbol\epsilon_\theta(x_t, t)$ is trained to predict the noise added at each step. By subtracting this estimate, we move one step toward a clean trajectory. This produces a learned Markov chain:
\begin{equation}
    p_\theta(\boldsymbol x_{t-1} \mid \boldsymbol x_t) = \mathcal{N}\!\left(\boldsymbol x_{t-1};  \boldsymbol \mu_\theta(\boldsymbol x_t, t), \sigma_t^2 \mathbf{I} \right),
    \label{eq:reverse}
\end{equation}
where the mean subtracts the estimated noise at step $t$, and the variance is the posterior variance following the derivation in \cite{ho_denoising_2020_final}:
\begin{equation}
\begin{aligned}
    \mu_\theta(\boldsymbol x_t, t)
    &= \frac{1}{\sqrt{1-\beta_t}}
    \left(
        \boldsymbol x_t - \frac{\beta_t}{\sqrt{1-\bar\alpha_t}}\, \boldsymbol \epsilon_\theta(\boldsymbol x_t, t)
    \right)
    \qquad
    \sigma_t^2 &= \tilde{\beta}_t
    \coloneqq \frac{1-\bar\alpha_{t-1}}{1-\bar\alpha_t}\,\beta_t.
\end{aligned}
\end{equation}
In the final generation step ($\boldsymbol x_{1} \longrightarrow \boldsymbol x_0$), the variance is set to zero.

\paragraph{Training objective.} We train our neural network $\boldsymbol\epsilon_\theta(x_t, t)$ by sampling a frame from the training dataset, applying the forward transform (Eq.~\ref{eq:forward}) at a timestep $t \sim \mathrm{Unif}\{1,\dots,T\}$, and optimizing its ability to recover the exact noise realization. For noise sampled as $\epsilon \sim \mathcal{N}(0,I)$, Ho et al.\ derive the following objective, which we minimize:

\begin{equation}
    \mathcal{L}(\theta)
    = \mathbb{E}_{x_0,\,\epsilon,\,t}
      \left[\, \lVert \epsilon - \epsilon_\theta(x_t,t) \rVert_2^2 \right] \quad \text{where~}   x_t = \sqrt{\bar\alpha_t}\, x_0 + \sqrt{1-\bar\alpha_t}\, \epsilon.
\end{equation}

\paragraph{Network architecture.} We implement $\boldsymbol\epsilon_\theta$ as a convolutional U-Net \cite{ronneberger_u-net_2015_final}: an encoder-decoder architecture that repeatedly halves spatial resolution while expanding channel depth, then mirrors the process with learned upsampling and skip connections so fine-scale details are combined with global context. Each block stacks ResNet convolutional residual layers \cite{he_deep_2016}, group normalization \cite{wu_group_2018_final}, and multi-head self-attention \cite{vaswani_attention_2023_final}. Sinusoidal timestep embeddings enter every block so the network always knows which diffusion step it is denoising. We trained both linear and cosine noise schedules with $T \in \{200, 500\}$ steps (we also tried 800, but these models did not converge within our time constraints), comparing four U-Net architectures: a compact 3-scale ResNet (1, 2, 4), a deeper 4-scale ResNet (1, 2, 2, 4), a ConvNeXt-inspired variant (1, 2, 2, 4; \cite{liu_convnet_2022_final}), and a 5-scale ResNet (1, 2, 4, 4, 8). The numbers in parentheses represent the channel multipliers across resolutions. Every variant shares the same training objective.

\subsection{Conditioning via inpainting with the CoPaint algorithm}
To produce forecasts, we use inpainting algorithms that modify the generation process so the final trajectory is conditioned on observed data. We first implemented RePaint \cite{lugmayr_repaint_2022_final}, a DDPM sampler that repeatedly overwrites the observed region during reverse diffusion (see Fig~\ref{fig:methods}c). We design a mask indicating which part of the image is to be imposed (the past for a forecasting task) or generated (the future). As in the generative process above, each iteration starts from the current sample $x_t$, but before denoising one step backward, we replace the imposed pixels with their ground-truth values noised to the corresponding time step $t$.
While RePaint shows impressive coherence in image-generation tasks and has been proven to generalize to unseen masks \cite{rout_theoretical_2023}, Rout et al.\ uncovered an alignment bias that leaves a visible discontinuity at the observed-forecast boundary, which affected Influpaint's performance during the 2022--2023 FluSight season.

CoPaint \cite{zhang_towards_2023_final} addresses this boundary mismatch by introducing a Bayesian formulation that jointly updates both revealed and unrevealed pixels at every denoising step, ensuring global coherence and preserving the diffusion trajectory. We apply CoPaint using its optimized DDIM (O-DDIM) sampler, an extension of the Denoising Diffusion \textit{Implicit} Model (DDIM) \cite{song_denoising_2022_final} that augments the deterministic reverse process with (i) a masked latent-refinement step at each time index and (ii) an optional time-travel schedule with re-noising that revisits earlier timesteps to reinforce the conditioning. These improvements lead to coherent inpainting and a smooth transition between masked and known regions. In our application, we find the transition between the observed (imposed) and generated parts of a trajectory to be much smoother with CoPaint. We refer the reader to S1 Appendix for our investigation of the inpainting schedule.

\subsection{Implementation details}
Influpaint was trained and evaluated on NVIDIA L40 and A100 GPUs. With a batch size of 512, Influpaint requires at least 20 GB of GPU memory. The iterative process used to inpaint a single forecast, for any mask, takes between 20 and 40 minutes to generate an ensemble of 512 conditioned trajectories. The framework is developed in Python using PyTorch, with implementations of CoPaint, RePaint, DDIM, and DDPM derived from their respective publications.

We relied on the hubverse \cite{hubs_coordinating_2025_final} to interact with other model forecasts, the \texttt{scoringutils} R package \cite{bosse_evaluating_2024} to score models, and the Delphi Epidata API \cite{farrow_delphi_2015} to fetch FluView data for construction of the Influpaint dataset.

\subsection{Statistical analysis}
For all forecasts submitted to FluSight or used in this work, Influpaint generates a predictive distribution represented by an ensemble of $n=512$ sampled trajectories. We converted these samples to the standard FluSight quantile format using 23 quantile levels: 0.01, 0.025, 0.05, 0.10, \ldots, 0.90, 0.95, 0.975, and 0.99. The same quantile grid was used throughout model evaluation. Forecast scores were computed in R with \texttt{scoringutils}, using the Weighted Interval Score (WIS), with lower values indicating better forecasts. Empirical interval coverage was also obtained from \texttt{scoringutils}; in the evaluation pipeline used for this study, the reported coverage metrics correspond to the 50\% and 90\% central prediction intervals. Coverage summaries reported in the manuscript are means of these per-forecast coverage indicators over the relevant set of forecast-observation pairs. For unconditional generation analyses, quantiles and envelopes were computed across $n=512$ generated trajectories. Pairwise cross-state weekly incidence correlations were calculated from state-level weekly hospitalization trajectories and summarized by their mean across state pairs.

\section*{Acknowledgments}
We thank the CDC and the organizers of the FluSight forecasting challenge for maintaining the open forecasting platform used in this study, as well as the Flu Scenario Modeling Hub coordinating team and participating modeling groups for providing access to simulated influenza trajectories. We also acknowledge the \textit{flepiMoP} teams for contributing the Round~1 trajectories that formed part of our training dataset. We thank Anton Geraschenko and Jessie K. Edwards for helpful advice.

\section*{Funding}
This work was supported by the National Institutes of Health (NIH, award \textit{5R01AI102939}). This project was made possible by cooperative agreement CDC-RFA-FT-23-0069 from the CDC's Center for Forecasting and Outbreak Analytics. Its contents are solely the responsibility of the authors and do not necessarily represent the official views of the Centers for Disease Control and Prevention.

\section*{Author contributions}
J. Lemaitre conceived the study, developed the methodology, curated the data, implemented the software, performed the analyses, generated the figures, and wrote the original draft. J. Lessler conceived the study, developed the methodology, interpreted the results, supervised the work, and substantively revised the manuscript. Both authors approved the final manuscript.

\section*{Competing interests}
J. Lemaitre reports consulting for Pfizer, Inc. J. Lessler declares no competing interests.

\section*{Data availability}
The pre-trained model weights used in this publication, along with all forecast trajectories, model rankings, model scores, and mask experiment results, are provided on Zenodo at \url{https://doi.org/10.5281/zenodo.22699980}. Influpaint real-time submissions are available in the CDC repositories \url{https://github.com/cdcepi/Flusight-forecast-data} for the 2022--2023 season (folder: \texttt{data-forecasts}) and \url{https://github.com/cdcepi/FluSight-forecast-hub} for the 2023--2024 and 2024--2025 seasons (folder: \texttt{model-output}), under the model name \texttt{UNC\_IDD-InfluPaint}.

\section*{Code availability}
The code used to train, run, and evaluate Influpaint is available on GitHub at \url{https://github.com/ACCIDDA/Influpaint}, and the version used for the present publication is archived on Zenodo at \url{https://doi.org/10.5281/zenodo.22712934}. A single command reproduces the figures presented in the main text and Supporting Information using the saved outputs in the Zenodo data repository (see above). A guide to reproducing the analyses and applying the Influpaint framework to other problems is available in the package documentation at \url{https://accidda.github.io/influpaint/}.

\section*{Supporting Information}
\noindent\textbf{S1 Appendix. Supplementary methods and results.}

\printbibliography

\clearpage
\begin{refsection}
\appendix
\setcounter{figure}{0}
\renewcommand{\thefigure}{\Alph{figure}}
\renewcommand{\theHfigure}{S\arabic{figure}}

\section*{S1 Appendix. Supplementary methods and results.}

\section{Supporting Information}
\subsection{Evaluation of competing model formulations}
To identify the factors that most influence Influpaint's forecasting performance, we conducted a series of one-at-a-time ablations around a reference configuration. The baseline used a 500-step cosine DDPM with a three-scale U-Net, the 30\% surveillance / 70\% simulated data mix, square-root scaling, no training-time enrichment, and the short-jump CoPaint schedule with no time travel. Forecasts were evaluated on 1--4-week horizons during the 2023--2024 and 2024--2025 FluSight seasons using the Weighted Interval Score (WIS).

\begin{figure} \centering
\includegraphics[height=3in]{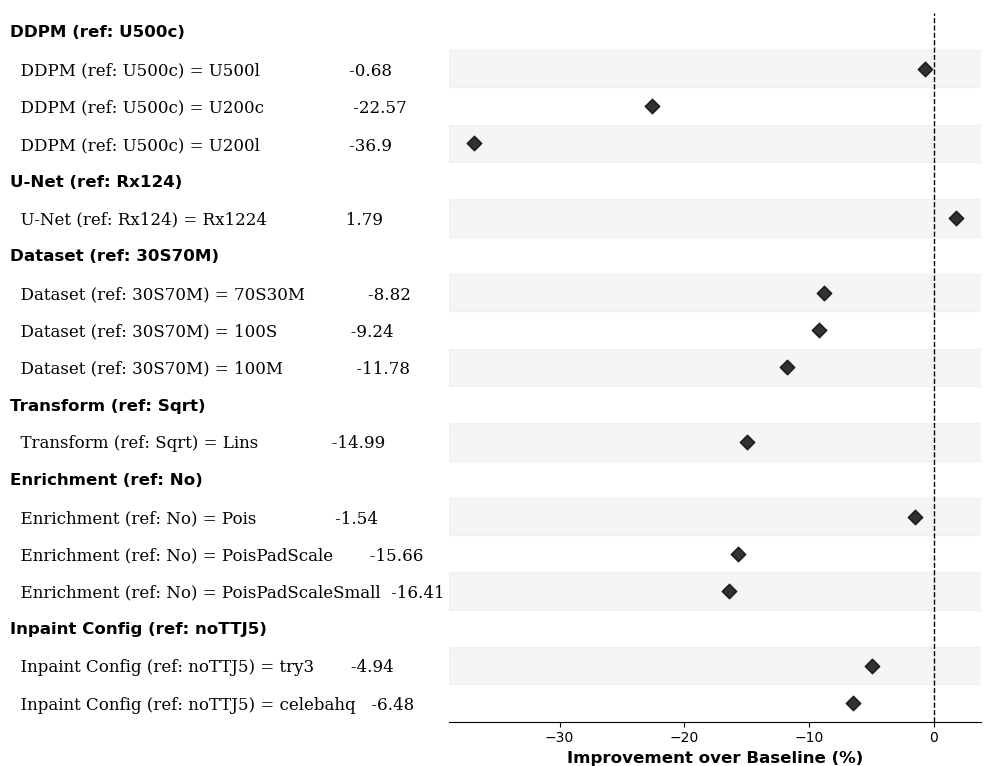}
\caption{\textbf{Ablation effects on WIS.} Each point shows the percentage improvement in summed absolute WIS for a model variant relative to the baseline configuration. Positive values indicate improved forecasting performance relative to the baseline, whereas negative values indicate worse performance. The first group reports diffusion-model denoising schedules, comparing 500 or 200 diffusion steps with either a linear (\textit{l}) or cosine (\textit{c}) variance schedule. The second group compares U-Net architectures, with \textit{124} denoting a three-scale architecture with channel multipliers $(1,2,4)$ and \textit{1224} a deeper four-scale architecture with channel multipliers $(1,2,2,4)$. The dataset group compares training sets denoted \textit{xxSyyM}, where \textit{xx}\% of samples come from surveillance data and \textit{yy}\% from modeled trajectories. The transform group compares square-root scaling of incidence values (\textit{Sqrt}) with linear scaling (\textit{Lins}). The enrichment group evaluates additional training perturbations: \textit{No} (none), \textit{Pois} (Poisson resampling), \textit{PoisPadScaleSmall} (Poisson resampling with narrow temporal padding and intensity scaling), and \textit{PoisPadScale} (Poisson resampling with wide temporal padding and intensity scaling). The final group compares the three inpainting configurations described in the main text.}
\label{fig:forest_effect}\end{figure}

Ablations varied diffusion depth and schedule, data composition, observation transforms, training enrichments, and inpainting schedules. We evaluated forecasting performance as the percent change in absolute WIS relative to the baseline (Fig~\ref{fig:forest_effect}).

The number of diffusion steps had the largest effect: models trained with T{=}500 diffusion steps outperformed those trained with shorter (T{=}200) schedules. The choice of observation transform had the next-largest effect, with square-root scaling yielding better performance than linear scaling. Training-time enrichment (Poisson resampling, temporal padding, or intensity scaling) degraded performance, suggesting that such perturbations disrupt the inductive bias learned from the training data. In contrast, architectural variants within the same capacity class and alternative CoPaint schedules had minimal impact (Fig~\ref{fig:forest_effect}).

\paragraph{Inpainting schedules evaluated.} We evaluate three CoPaint/ODDIM sampling schedules that differ only in how strongly and how often the observed data are enforced along the reverse diffusion trajectory. \textit{Time travel} means re-noising the current sample to a higher diffusion time and repeating the conditioned denoising process. When time travel is enabled, the \textit{jump length} \(J\) defines the number of diffusion steps for each return. A \textit{latent-refinement step} reduces the discrepancy between the predicted trajectory and the observed data. We compare the following schedules: (i) a short jump (\(J{=}5\)) with time travel and five latent-refinement steps per time index (\texttt{celebahq\_try3}); (ii) the same jump length without time travel and with two latent-refinement steps (\texttt{celebahq\_noTTJ5}); and (iii) a longer jump (\(J{=}10\)) with time travel and two latent-refinement steps (\texttt{celebahq}).
Because conditioning is performed entirely at prediction time (the diffusion model is trained unconditionally), any surveillance/forecast mask can be used without retraining; this is consistent with theoretical results on resampling-based diffusion inpainting generalization \cite{rout_theoretical_2023}. The data used to evaluate predictive performance were not used for model training and are introduced to Influpaint only at generation time.

\subsection{Relationship between training loss and forecasting performance}

\begin{figure} \centering
\includegraphics[height=3in]{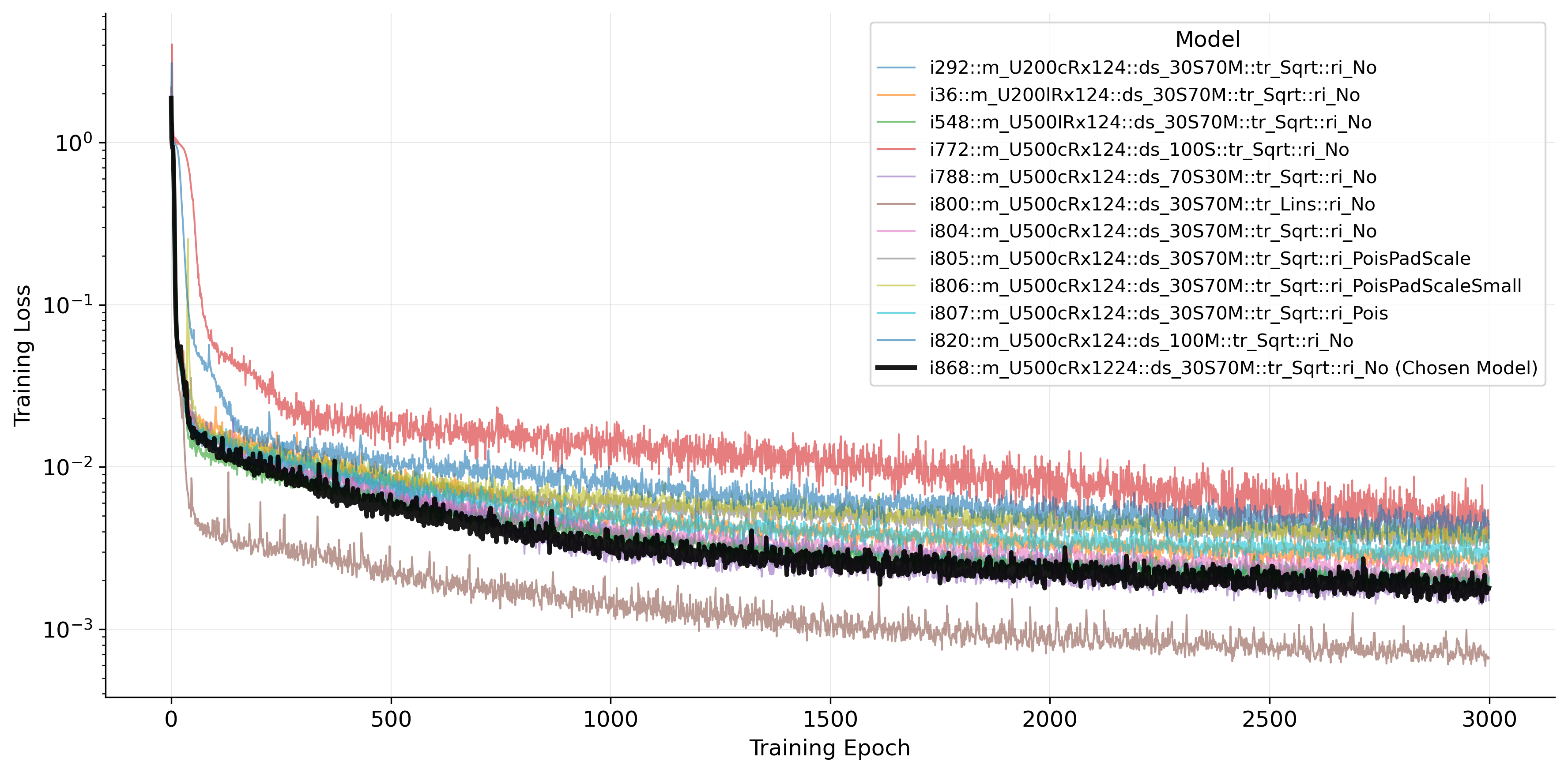}
\caption{\textbf{Training losses during Influpaint calibration.} Curves show the training loss over 3,000 epochs for the candidate model configurations evaluated during calibration. The model selected for downstream analyses, based on its combined ranking in relative and absolute WIS, is highlighted in black.}
\label{fig:training_loss}\end{figure}

\begin{figure} \centering
\includegraphics[height=2.5in]{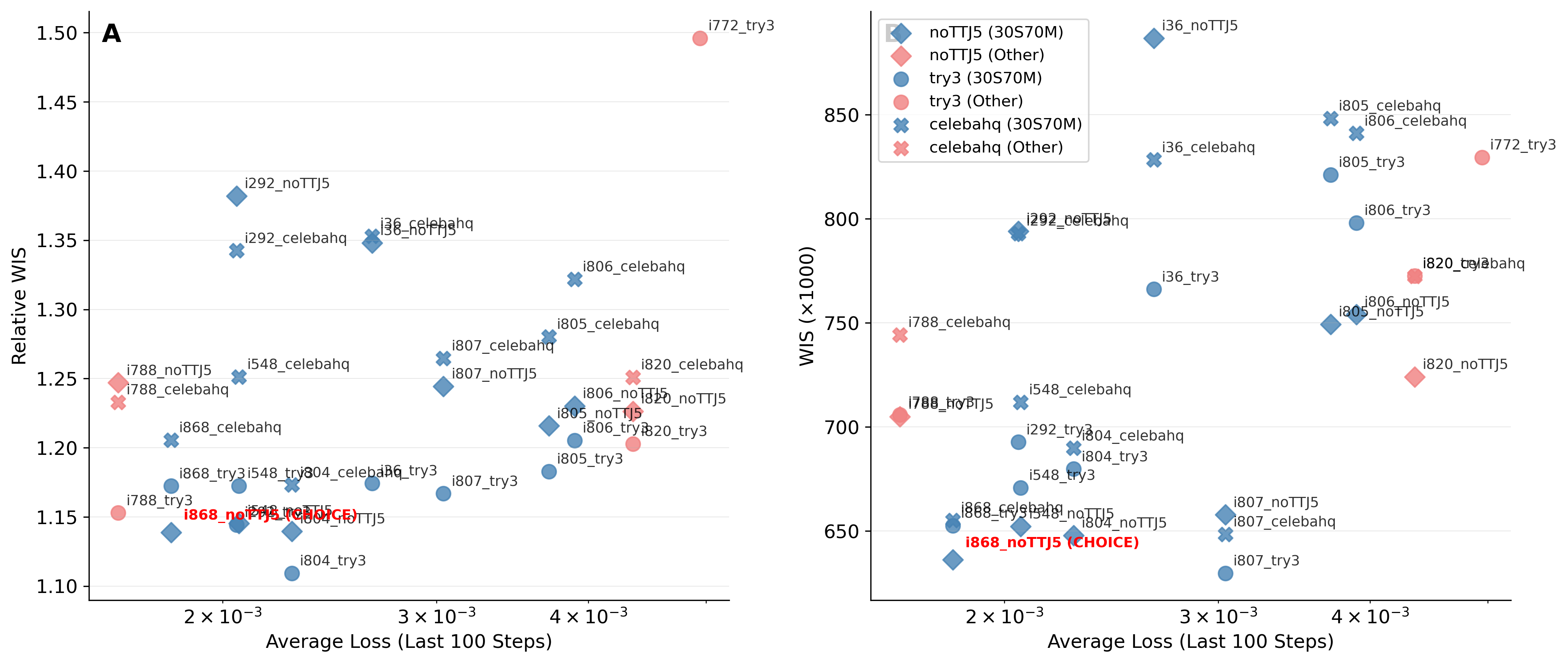}
\caption{\textbf{Relationship between training loss and forecasting performance.} Final training loss is compared with relative WIS (\textbf{a}) and absolute WIS (\textbf{b}) across candidate model configurations. Each point represents one model variant, and the configuration selected for downstream analyses is highlighted in red.}
\label{fig:lossNwis}\end{figure}

During training, Influpaint's loss (Fig~\ref{fig:training_loss}) measures performance on unconditional generation rather than directly measuring forecasting performance. When we compared this loss with forecasting skill, we found a positive correlation, though not a strict ordering across models (Fig~\ref{fig:lossNwis}). This suggests that generative fit is informative about forecast quality, but does not fully determine it. Directly training diffusion models for forecasting may therefore improve performance further. Even so, these results show that training for unconditional generation already yields models with substantial forecasting skill.

\subsection{Realized FluSight forecasts}
\begin{figure} \centering
\includegraphics[width=\textwidth]{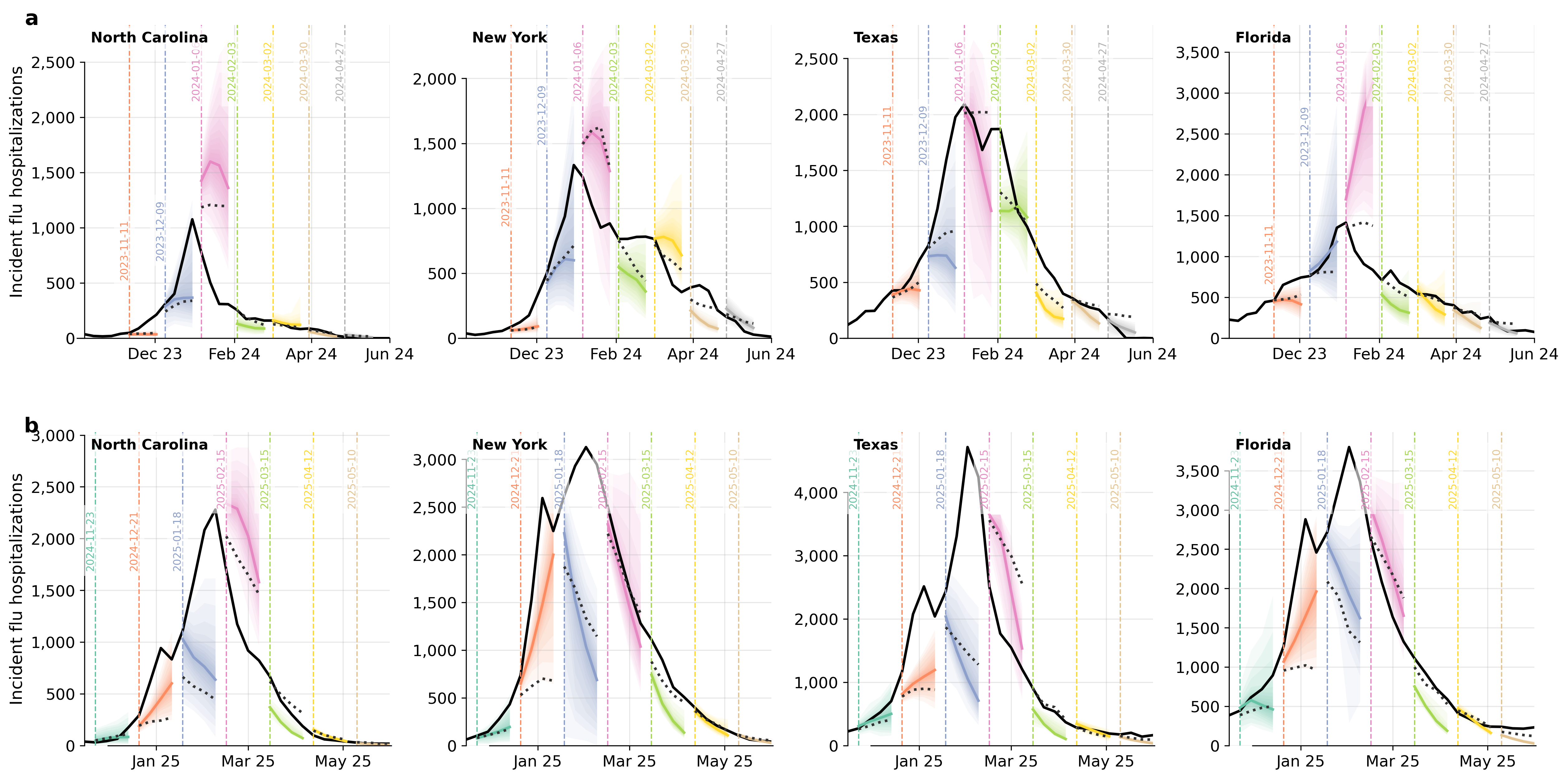}
\caption{\textbf{a.} Each panel shows submitted 4-week-ahead FluSight forecasts from \texttt{UNC\_IDD-InfluPaint} for North Carolina, New York, Texas, and Florida at the same forecast dates used in \textbf{Figure 2} of the main text (colored dashed vertical lines). Forecast uncertainty is summarized by the submitted quantiles (colored fan) and median (colored line). The solid black line shows the observed final values. For the same forecast dates, the FluSight ensemble forecast is shown as a dotted line.
\textbf{b.} Same as panel \textbf{a}, but for the 2024--2025 influenza season.}\label{fig:realized}\end{figure}

We show a sample of real-time Influpaint forecasts submitted to the FluSight hub in Fig~\ref{fig:realized}.

\FloatBarrier

\printbibliography

\end{refsection}

\end{document}